\documentclass{article} 
\usepackage{iclr2019_conference,times}

\usepackage{amsmath,amsfonts,bm}

\def\eqref#1{equation~\ref{#1}}

\def\1{\bm{1}}

\DeclareMathAlphabet{\mathsfit}{\encodingdefault}{\sfdefault}{m}{sl}
\SetMathAlphabet{\mathsfit}{bold}{\encodingdefault}{\sfdefault}{bx}{n}

\usepackage{hyperref}
\usepackage{url}
\usepackage{graphicx} 
\usepackage{caption} 
\usepackage{subcaption} 
\usepackage{booktabs} 

\title{Adversarial Attacks for Optical Flow-Based Action Recognition Classifiers}

\author{Nathan Inkawhich, Matthew Inkawhich, Yiran Chen \& Hai Li \\
Department of Electrical and Computer Engineering\\
Duke University\\
Durham, NC 27701, USA \\
\texttt{\{nathan.inkawhich,matthew.inkawhich,yiran.chen,hai.li\}@duke.edu}
}

\iclrfinalcopy 
\begin{document}

\maketitle

\begin{abstract}

The success of deep learning research has catapulted deep models into production systems that our society is becoming increasingly dependent on, especially in the image and video domains. However, recent work has shown that these largely uninterpretable models exhibit glaring security vulnerabilities in the presence of an adversary. In this work, we develop a powerful untargeted adversarial attack for action recognition systems in both white-box and black-box settings. Action recognition models differ from image-classification models in that their inputs contain a temporal dimension, which we explicitly target in the attack. Drawing inspiration from image classifier attacks, we create new attacks which achieve state-of-the-art success rates on a two-stream classifier trained on the UCF-101 dataset. We find that our attacks can significantly degrade a model's performance with sparsely and imperceptibly perturbed examples. We also demonstrate the transferability of our attacks to black-box action recognition systems.
%
%
%
\end{abstract}

\section{Introduction}

As machine learning (ML) for computer vision becomes more popular, and ML systems are integrated into production level technologies, the security of ML models becomes a serious concern. Until recently, researchers have been focused on pushing the boundaries of model accuracy, while not prioritizing model security as a first-order design constraint. Many previous works \citep{FGSM,CW,Deepfool,JSMA,LBFGS} have shown that adding imperceptible noise to an image can easily fool a CNN-based classifier, but no such work has been done in the video and action recognition domain. A unique property of the action recognition domain is the presence of a temporal dimension, which has been shown to be crucially important to the efficacy of current classifiers \citep{IntegrationOflow,2stream,QuoVadis}. It also provides a new axis to attack. 
%
%

In this work, we consider the two-stream architecture \citep{2stream} and its variants, which represent the market share of the current state-of-the-art action recognition models. We isolate the motion stream classifier of a two-stream model as our model to attack. Previous work has shown that the standalone motion stream, trained on stacks of optical flow fields, outperforms the standalone spatial stream, and is not significantly worse than the full model's capability \citep{2stream,ConvFusion,LongTemporal,STMultiplier}. Therefore, we contend that if the temporal stream is compromised, the integrity of the entire classifier is compromised. Also, any image domain attack may be applied to the spatial classifier. 

The traditional variational optical flow calculation algorithms are non-differentiable, meaning that performing a known white-box image attack to directly perturb the video frames is not possible due to lack of gradient flow. To circumvent this, we consider the FlowNet 2.0 model, which is a convolutional neural network that estimates optical flow fields \citep{Flownet2}. If we consider a target model that is a combination of the FlowNet 2.0 model and the temporal stream CNN, we have an end-to-end differentiable model that takes a raw video and returns a classification. We leverage this composite model to obtain gradients of the \textit{temporal stream} classification loss with respect to the \textit{spatial} input video frames, which is precisely what is needed for an effective white-box attack. We then demonstrate the effectiveness of our white-box adversarial examples on black-box models trained with variational optical flow algorithms. The observed transferability of our attack on black-box models greatly increases its practical usefulness in the real-world.

Overall, our contributions are as follows:
\begin{enumerate}
\item We create a white-box, untargeted attack for a two-stream action recognition classifier;
\item We show that the performance of action recognition systems can be completely degraded with sparsely and imperceptibly perturbed examples;
\item We create a black-box attack to produce examples that can transfer to other action recognition systems with alternative optical flow and CNN algorithms;
\item We introduce the idea of salient video frames in the context of video classification.
\end{enumerate}

\section{Related Work}

\textbf{Action Recognition.}
There are several general methods for modeling motion in action recognition classifiers. One approach is to use 3D-Convolution on stacked spatial frames, attempting to learn spatial features and temporal difference features concurrently \citep{3DConv,BoW}. Another strategy involves training LSTM models on sequences of features extracted with convolutional layers of image CNNs \citep{VidLSTM,CNN-LSTM}. Finally, a host of research efforts train CNN classifiers on optical flow displacement fields and spatial frames separately \citep{2stream,ConvFusion,TSN,BSS,STResidual,STMultiplier,QuoVadis,ActionVLAD,AsyncTF,IntegrationOflow,LongTemporal}. Here, we will focus on these optical flow-based systems, as they represent a significant portion of the most effective methods to date \citep{QuoVadis,IntegrationOflow}. The first design to use CNNs with optical flow is the two-stream model \cite{2stream}. This model uses two CNNs: one trained on spatial frames and the other trained on stacks of sequential optical flow displacement fields. To produce a final prediction, each stream makes a classification independently, and the predictions are fused. Since this innovation, there have been many works that employ the idea of a two-stream architecture and use optical flow specifically to model motion. \cite{ConvFusion} propose convolutional network fusion, where spatial stream and temporal stream feature maps are combined before making a prediction. \cite{STResidual} and \cite{STMultiplier} describe spatio-temporal multiplier networks and spatio-temporal residual networks, which leverage deep residual networks for both streams by allowing the streams to share information via the residual connections. Finally, one of the top performing methods for action recognition is \cite{QuoVadis} Two-Stream Inflated 3D ConvNet (I3D) architecture, which combines techniques from the two-stream model, 3D convolution, transfer learning, and the Inception model. Each of the aforementioned methods are considered state-of-the-art, and all involve a two-stream model where the motion stream operates on optical flow. This distinction serves as motivation for the framing of our attack model.

\textbf{Optical Flow.}
Optical flow generation consists of two broad categories: variational and deep learning-based. Two common variational techniques are Farneback \citep{Farneback} and TV-L1 \citep{TVL1,TVL12}. The Farneback algorithm estimates frame neighborhoods by quadratic polynomials using the polynomial expansion transform, and is optimized using a coarse-to-fine strategy. The TV-L1 algorithm is a more recent approach that works to minimize a function containing a data fidelity term using the $L_1$ norm and a regularization term based on the total variation of the flow \citep{TVL1}. TV-L1 is more accurate, and shows increased robustness against illumination changes, noise, and occlusion. Recent works also show that deep convolutional neural networks trained in a supervised fashion can be a fast and effective way to estimate optical flow \citep{Flownet,Flownet2}. Specifically, FlowNet2 has been shown to be as accurate as state-of-the-art variational optical flow estimation methods, while running significantly faster.

\textbf{Adversarial Attacks.}
Crafting adversarial examples has become an increasingly popular area of research recently, especially for image classifiers. Roughly speaking, the space is divided into two general categories: white-box and black-box. White-box attacks assume full knowledge of the model and parameters, and often use gradient information in the attack \citep{FGSM,CW,Deepfool,JSMA,LBFGS}. Black-box attacks on the other hand treat the model as an oracle and do not have intimate knowledge of the model architecture or specific parameters \citep{Transferability}. Perhaps the most popular white-box attack method is the Fast Gradient Sign Method (FGSM) \citep{FGSM}. FGSM uses the gradient of the loss w.r.t. the input to adjust the image in the direction that maximizes the loss. This results in an imperceptible noise field being added to the original image that significantly degrades the classification performance of the model. Since, attacks such as Carlini-Wagner Attack \citep{CW}, Jacobian Saliency Map Attack (JSMA) \citep{JSMA}, Deepfool \citep{Deepfool}, and Iterative Least Likely \citep{it-FGSM}, have leveraged the gradient information of the model in some fashion to create adversarial examples. Another important finding from these attacks is that an adversarial example computed from one model is often adversarial to other models. This principle is called transferability and has been studied extensively for image classification systems \citep{Transferability,transfer2,transfer3}.

\section{Attack Methodology}

\subsection{Model Under Attack}

\begin{figure}[h]
  \includegraphics[width=.9\columnwidth]{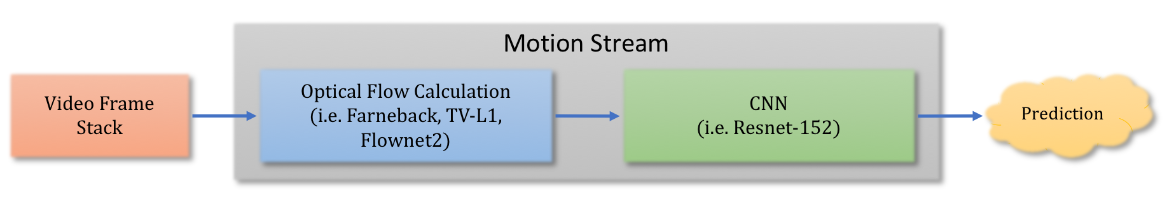}
  \centering
  \caption{Model under attack, representing the motion stream of a two-stream action recognition classifier. The model inputs a stack of video frames, internally calculates and classifies an optical flow stack, then outputs a prediction.}
  \label{fig:motion_stream}
\end{figure}

In this work we define the Model Under Attack (MUA) as the isolated motion stream of the two-stream architecture, where motion is modeled as optical flow fields. Fig. \ref{fig:motion_stream} shows a depiction of this general model configuration. Note, the input is a stack (usually length 11) of video frames, optical flow is calculated internally, and the output is the classifier's prediction.

This MUA represents any model that uses optical flow stacks to model motion. Previous works \citep{2stream,ConvFusion,LongTemporal,STMultiplier} report that the standalone motion stream of a two-stream model outperforms the standalone spatial stream. On the UCF101 dataset, \cite{2stream} show that the spatial stream of a two-stream model alone achieves 73.0\% accuracy, while the temporal stream alone achieves 83.7\%. Thus, if the motion stream can be fooled, the entire model is compromised.

\subsection{Attack Setup}

\begin{figure*}[!hbt]
  \includegraphics[width=\textwidth]{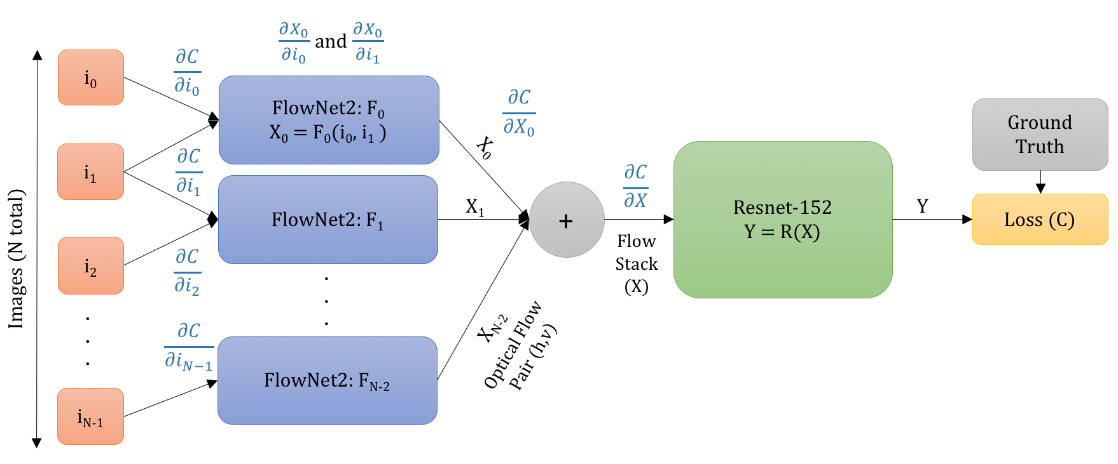}
  \centering
  \caption{Detailed diagram of the MUA. Text in black relates to the forward pass and text in blue relates to the backward pass. Optical flow is calculated between frame pairs and concatenated to form an optical flow stack which is classified with the CNN.}
  \label{fig:full_sys}
\end{figure*}

We define the input frame stack as $\boldsymbol{i}$, where $\boldsymbol{i}$ is a series of $N$ discrete frames, i.e. $\boldsymbol{i} = \left[ i_{0}, i_{1}, i_{2}, \ldots , i_{N-1} \right]$. The optical flow calculation that produces a single displacement field between two successive frames is $X_{n} = F_{n}(i_{n}, i_{n+1})$, where $X_{n}$ is the horizontal and vertical displacement fields, and $F_{n}$ is the optical flow function (i.e. Farneback, TV-L1, FlowNet2) that operates on frames $i_{n}$ and $i_{n+1}$. Since action recognition systems operate on stacks of optical flows, we define $\boldsymbol{X}$ as an optical flow stack, which is formed by concatenating the individual flow fields $X_{0}, X_{1}, X_{2}, \ldots, X_{N-2}$ while maintaining the respective ordering in time. For convenience, let $F(\boldsymbol{i})=\boldsymbol{X}$ represent an entire optical flow stack given a video frame stack ($\boldsymbol{i}$) as input. The CNN classifier $R$, with loss function $C$, is a function of the optical flow stack, and $R(\boldsymbol{X})$ (or $R(F(\boldsymbol{i}))$) is a softmax array of probabilities where the class prediction is $argmax(R(\boldsymbol{X}))$. The goal of the attack is to apply the least amount of noise ($\epsilon$), s.t. $argmax(R(F(\boldsymbol{i}))) \neq argmax(R(F(\boldsymbol{i}+\epsilon)))$ assuming the initial prediction is correct. In this case we define the perturbed video, $\boldsymbol{i}' = \boldsymbol{i}+\epsilon$, to be adversarial. 

The proposed white-box attack draws from the ideas of FGSM \citep{FGSM} and iterative FGSM \citep{it-FGSM}. FGSM attacks adjust the input image using a scaled version of the sign of the gradient of the loss w.r.t. the input. In this context, the FGSM method only calculates the gradients through the classifier CNN, w.r.t. the optical flow stack ($\partial C / \partial \boldsymbol{X}$). This means it does not provide the information necessary to adjust the input video frames ($\boldsymbol{i}$) directly, as the optical flow calculation is performed internally within the model. For an action recognition attack, we must compute gradients through the classifier \textit{and} the optical flow calculation. One problem when using variational algorithms such as Farneback and TV-L1 is that the optical flow calculation is non-trivially differentiable. Therefore, we use FlowNet2, a convolutional neural network that estimates optical flow between successive frames. Using FlowNet2, we can easily compute the gradients through the model ($F_n$), and define the derivative of the output w.r.t. the inputs of FlowNet2 as $\partial X_n / \partial i_n$ and $\partial X_{n} / \partial i_{n+1}$. With chain rule and $\partial C / \partial \boldsymbol{X}$ we can compute the gradient of the loss w.r.t each input image, for $n$ in $[0,N)$ as

\begin{equation} 
	\label{eq:dC_di}
    \frac{\partial C}{\partial i_n} =
	\begin{cases}
        \frac{\partial C}{\partial X_n} \frac{\partial X_n}{\partial i_n} & n = 0 \\[8pt]
		\frac{\partial C}{\partial X_{n-1}}\frac{\partial X_{n-1}}{\partial i_n} + \frac{\partial C}{\partial X_n}\frac{\partial X_n}{\partial i_n} & 0<n<N-1 \\[8pt] 
        \frac{\partial C}{\partial X_{n-1}}\frac{\partial X_{n-1}}{\partial i_n} & n=N-1. \\
	\end{cases} 
\end{equation}

As a result of (\ref{eq:dC_di}), we obtain the gradient of the loss w.r.t. the input images themselves. Fig. \ref{fig:full_sys} shows the diagram of the model used in the attack. The video frame stack is represented as separate images, there is one FlowNet2 model for each pair of sequential frames, and the individual optical flow displacement fields are concatenated before being input into the CNN classifier. The text shown in black represents the signals that are present in the forward pass and the text in blue pertains to the backward pass. The following sections are dedicated to describing each variant of our attack, and how they use the information computed in the backward pass to create adversarial examples.

\subsection{One Shot Attack}

The baseline attack variant is the \textbf{\textit{one-shot}} attack, which only involves one forward and one backward pass. Recall, we have calculated the partial derivatives of the cost w.r.t. each input frame ($\partial C / \partial i_n$), which can be written in terms of the gradient w.r.t the video as

\begin{equation} \label{eq:gradient_C}
\nabla C( \boldsymbol{\theta}, \mathbf{i}, y) = \bigg[ \frac{\partial C}{\partial i_0}, \frac{\partial C}{\partial i_1}, \ldots, \frac{\partial C}{\partial i_{N-1}} \bigg] ^T 
\end{equation}

where $\boldsymbol{\theta}$ represents the model parameters, $\boldsymbol{i}$ is the video frame stack, and $y$ is the ground truth label of $\boldsymbol{i}$. We then update all of the input images as follows

\begin{equation} \label{eq:img_update}
i_n' = i_n + \epsilon * sign(\nabla C_{i_{n}}( \boldsymbol{\theta}, \boldsymbol{i}, y)). 
\end{equation}

As a result of the one-shot update, all of the images in the input video are perturbed at all locations by a small amount ($\epsilon$), in the direction that will maximize the loss. To maintain the original distribution of the input, any pixel values that exceed the original range of $[0,255]$ are clipped to fit the range. Notice, the \textit{one-shot} attack is time and computation efficient because only one forward pass and one backward pass must be computed. However, it is limited in the sense that it perturbs every pixel of every frame and has no sparsity constraint. Ideally, the perturbations would be sparse in time, meaning not all frames would be perturbed. We can achieve sparsity through iteration, which is the goal of the iterative attacks. 

\subsection{Iterative Attacks}

The \textbf{\textit{iterative-saliency}} attack variant is an attempt at crafting adversarial examples without perturbing all frames. The idea is to iteratively perturb the video using (\ref{eq:img_update}), one frame at a time. Frames are perturbed in order of decreasing saliency, where frame $i_n$'s saliency $S_{i_n}$ is defined as

\begin{equation} \label{eq:calc_saliency}
 S_{i_n} = \frac{1}{HW} \sum_{h=0}^{H-1} \sum_{w=0}^{W-1} \bigl | \nabla C_{i_n}( \boldsymbol{\theta}, \mathbf{i}, y) [h][w] \bigr |
\end{equation}

Here, $H$ and $W$ represent the spatial height and width of the video frames, respectively. In other words, this scalar quantity of saliency is the average magnitude of the gradient w.r.t. a single frame. This notion of saliency is inspired by \cite{saliency}, which states that ``[an interpretation of] image-specific class saliency using the class score derivative is that the magnitude of the derivative indicates which pixels need to be changed the least to affect the class score the most.'' Therefore, the intuition behind the attack is that by iteratively perturbing the frames of a video that the prediction is most sensitive to (i.e. have high average saliency as computed by (\ref{eq:calc_saliency})), we can craft sparsely perturbed adversarial examples.

At each iteration, we perform a full forward pass to check if the video frame stack is adversarial (i.e. calculate $R(F(\boldsymbol{i}))$). If the video is not adversarial, we have two options: perturb the next most salient frame using the original gradient information calculated during the first forward pass, or recalculate $\nabla C( \boldsymbol{\theta}, \mathbf{i}, y)$ and $S_{i}$ before perturbing the next frame. For this reason, we create two variants of the \textit{iterative-saliency} attack: \textbf{\textit{iterative-saliency}} and \textbf{\textit{iterative-saliency-RG}} (RG for refresh gradient). Put more explicitly, the \textit{iterative-saliency} variant calculates the gradient ($\nabla C( \mathbf{ \theta }, \mathbf{i}, y)$) and saliency values for each frame once after the initial forward pass. It then continues iteratively perturbing frames with (\ref{eq:img_update}) until either a misclassification is reached or all frames have been perturbed. The \textit{iterative-saliency-RG} variant recalculates the gradients and the saliency values for all frames after every unsuccessful iteration. However, it is not allowed to perturb the same frame twice.

The appeal of the \textit{iterative-saliency} variant is that it is less expensive, as the backward pass only needs to be computed once, w.r.t. the original video. The RG variant requires a backward pass for every forward pass, but the perturbations are better optimized to the changing frames. If all frames are perturbed and the video is still correctly classified, the attack has failed.

\section{Implementation Details}

\textbf{Dataset.} For our experiments we use the UCF-101 dataset \cite{ucf101}, which is among the most common action recognition and video classification benchmarks and is tested in all action recognition related works. The dataset consists of 13,320 videos from 101 human action categories such as Archery, Baseball Pitch, Playing Violin, Typing, etc. The videos have been collected from YouTube and have an average duration of 7.2 seconds. Each video clip has a uniform frame rate of 25 fps with spatial size 320x240 pixels. In this work specifically, we adhere to the official split-01 for reporting training and testing results. We also subsample the videos to 12 fps for convenience.  

\textbf{Classifier Setup.} There are many hyper-parameters of action recognition models, including spatial size and channel depth of input tensors. In this work we use optical-flow-stack classifiers with a CHW input volume of (20,224,224). In accordance with \cite{2stream}, we define optical-flow-stack length to be 10. To achieve this, we extract contiguous sets of 11 frames from each video, calculate the optical flow fields (horizontal and vertical flow pair) between each pair of adjacent frames, then stack the individual pairs depth-wise to achieve depth 20. This follows the optical flow stacking technique used for the original two-stream architecture.

\textbf{Training.} There are several deep learning models that must be trained for this research, including the FlowNet2 optical flow model and separate classifiers for each optical flow method. Before any training, we generate TV-L1 \citep{TVL1} and Farneback \citep{Farneback} optical flow datasets using the OpenCV implementations with default parameters. An optical flow pair is calculated between each adjacent frame of all videos and the horizontal and vertical fields are separately saved as gray-scale PNG images. To maintain resolution, the optical flow fields are clipped to $[-20,20]$. 

Next, we fine-tune a FlowNet2 \citep{Flownet2} model using NVIDIA's FlowNet2 PyTorch implementation \cite{flownet2-pytorch}. We fine-tune for 6 epochs using the previously generated TV-L1 optical flows as the ground truth. After fine-tuning, we generate the full optical flow dataset and save each flow pair to disk as a grayscale PNG with the same range. Now, we have three separate optical flow datasets for the TV-L1, Farneback, and FlowNet2 methods.

Many models have been suggested for use in the two-stream framework. Inspired by results from \cite{STResidual,STMultiplier}, we use a Resnet-152 \citep{Resnet} model as our CNN in the MUA. We train three Resnet-152 models, one for each optical flow dataset. Since the spatial frames are larger than the 224x224 spatial input of the CNN, during training we using random scaling and cropping data augmentations. However, during testing we use a simple center-crop for prediction. As a result of training for several hundred thousand iterations each, we have three similarly performing models. The CNNs trained on TV-L1, Farneback, and FlowNet2 optical-flow-stacks have split-01 stack-level test accuracies of 70.72\%, 68.94\%, and 74.01\%, respectively. Note, these are not the video level results reported in related papers. These baseline stack-level accuracies will serve as the baseline, for which we will attempt to degrade with our attacks.  It also shows that models trained on all three methods of optical flow yield similar results, so any of the methods may be a viable option for use in an action recognition system, depending on the application's requirements for speed and computational complexity.

\textbf{Stack Level vs. Video Level.} Before continuing, it is important to emphasize the difference between stack-level and video-level. As mentioned, the primary attack operates on the stack-level as this is the granularity that action recognition classifiers work. A stack refers to a set of 11 contiguous frames that have been sampled from a full length video. If the video is longer than 11 frames, then it potentially contains more than one stack, depending on sampling scheme. This stack of 11 frames is then used to create the length 10 optical-flow-stack that is fed to the classifiers. Video-level predictions refer to the practice of averaging stack-level predictions into a single prediction, which will be discussed further in Section 5.2.

\section{Experimental Results}

\subsection{Stack Level Results}

\begin{figure*}[!hbt]
  \includegraphics[width=\textwidth]{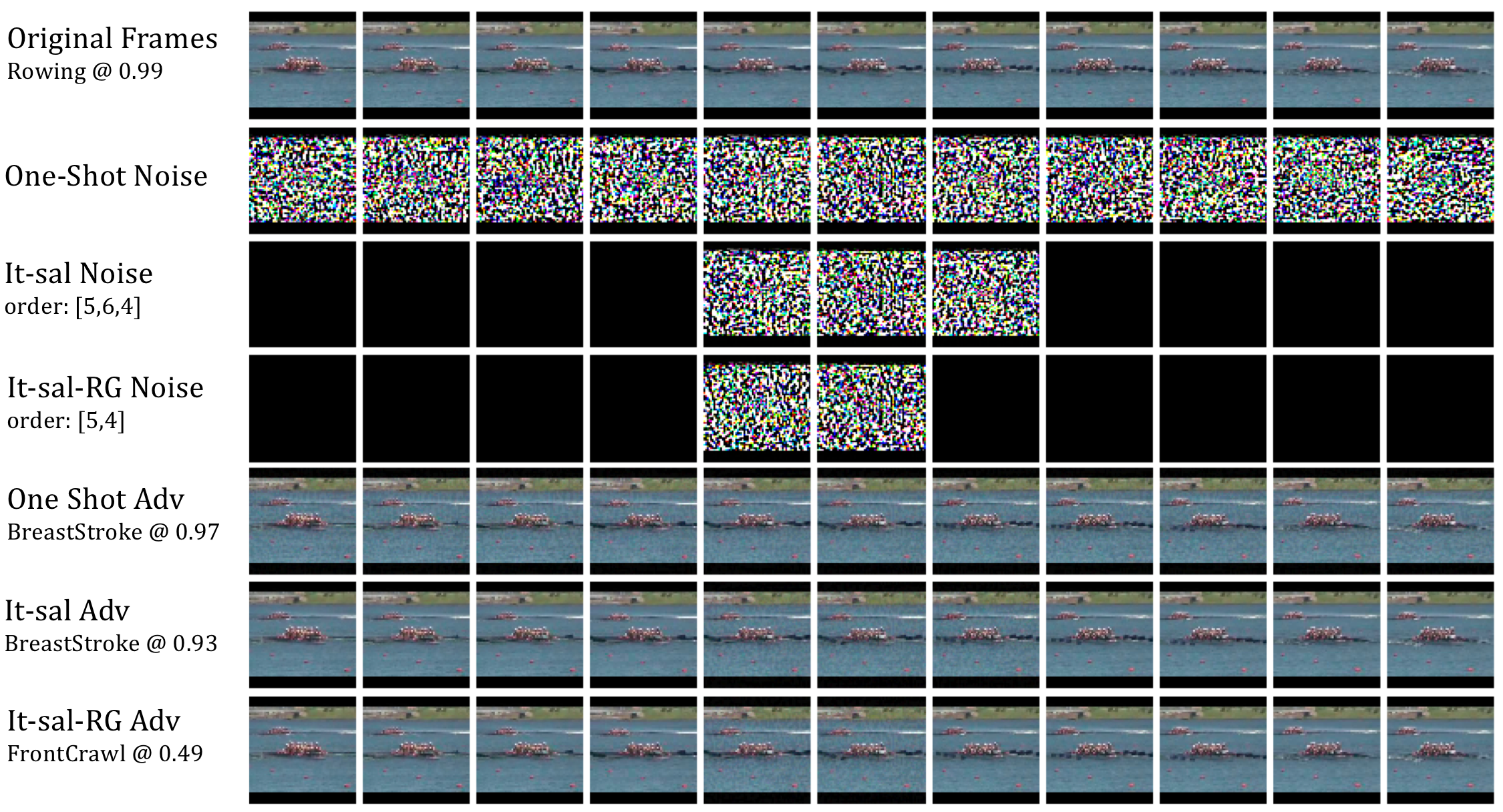}
  \centering
  \caption{Visualization of the perturbations required from each attack to cause a misclassification for a sample video clip. The adversarial examples shown are with $\epsilon = 0.025$.}
  \label{fig:perturbations}
\end{figure*}

The first result, shown in Fig.\ref{fig:perturbations}, is to visualize the three attacks at $\epsilon=0.025$, on a single stack. This image shows the fundamental differences between the attack variants, and also shows examples of perturbed frames. The original stack is classified as Rowing at 99.79\% confidence. The \textit{one-shot} attack perturbs all frames to cause a misclassification of BreastStroke at 97.88\% confidence. The \textit{it-saliency} attack perturbs 3 frames total (frame 5, then 6, then 4) and causes a misclassification of BreastStroke at 93.12\% confidence. The \textit{it-saliency-RG} attack perturbs 2 frames total (frame 5, then 4) and causes a misclassification of FrontCrawl at 49.58\% confidence. Interestingly, the perturbed classes all have to do with water which may indicate that water leaves a distinct signature in the optical flow. Finally, notice that the perturbations are almost indistinguishable even at this relatively high epsilon. For examples of perturbed frames at all of the tested epsilon values, see Fig. \ref{fig:perturbation_examples} in the Appendix.

\subsubsection{Accuracy Results}

The next result is the accuracy versus epsilon test for the stack-level classifier. Here, we sweep $\epsilon$ from 0 to 0.035 in steps of 0.005. At $\epsilon=0$, the model is not under attack and this represents the base top-1 accuracy. Intuitively, we would expect that as $\epsilon$ increases (i.e. the strength of the attack increases), accuracy monotonically decreases. Keep in mind that since this is a 101 class dataset, random accuracy is about 1\%. 

\begin{figure}[h]
\centering
\begin{subfigure}{.45\textwidth}
  \centering
  \includegraphics[width=\linewidth]{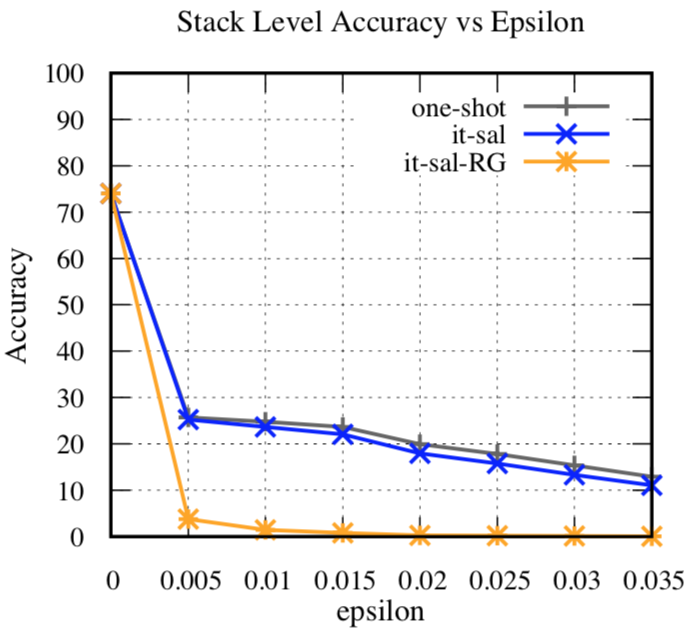}
  \caption{}
  \label{fig:stack_results_plot}
\end{subfigure}%
\begin{subfigure}{.45\textwidth}
  \centering
  \includegraphics[width=\linewidth]{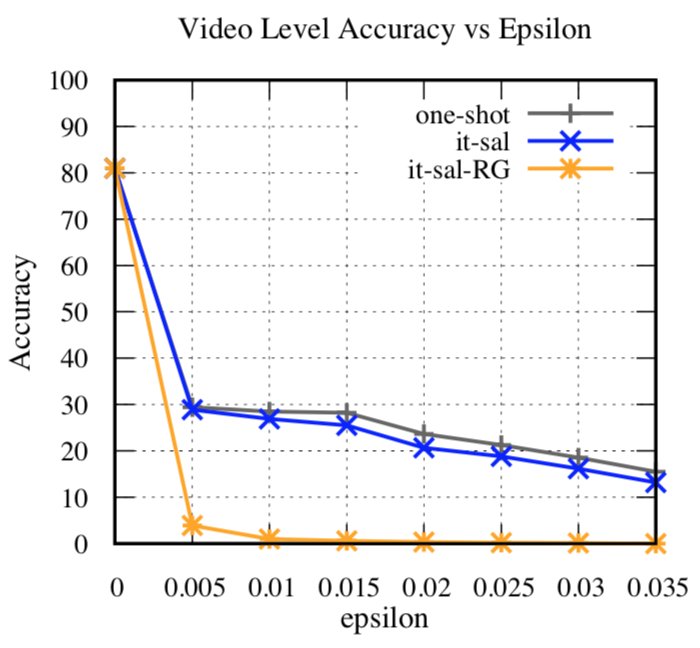}
  \caption{}
  \label{fig:video_results_plot}
\end{subfigure}
\caption{Plots showing how the accuracy of the classifier changes as epsilon changes for the three attack variants. Specifically, (a) shows how the stack-level accuracy changes and (b) shows the video-level accuracy.}
\end{figure}

Fig. \ref{fig:stack_results_plot} and the"Stk-A" columns of Table \ref{table:acc_results_table} show the stack level accuracy versus epsilon results for the attacks. From Fig. \ref{fig:stack_results_plot}, it is clear that as epsilon increases, accuracy consistently decreases for all three attacks. Also, we observe that the \textit{one-shot} and \textit{it-saliency} attacks perform very similarly, while the \textit{it-saliency-RG} attack is by far the most powerful. This result is sensible, as the two less powerful attacks use the same gradient information calculated in the first backward pass, while the RG attack is constantly updating the gradients at each step. One potential reason the \textit{it-saliency} slightly outperforms the \textit{one-shot} attack, is due to the unintended effects of adding and removing the noise field from frame to frame. The margin by which the RG attack outperforms the others is also significant, lowering the accuracy by about 22\% more at the weakest attack strength. The RG attack is also the only variant to achieve random accuracy, at $\epsilon=0.015$. It is also worth noting that the elbow in the curves appear at $\epsilon=0.005$, the weakest tested attack strength. There is a large drop in accuracy at this value, which is not matched at any other strength step. This may mean that there is a large contingent of data that lie near the decision boundaries that are easily adversarially perturbed. Most other examples lie a large distance away from the boundaries with not many data in-between.

\begin{table}[h]
  \centering
  \caption{Summary of stack and video level attack results for white-box attack}
  \label{table:acc_results_table}
  \begin{tabular}{@{}cccccccccc@{}}
  & \multicolumn{3}{c}{\textit{one-shot}} & \multicolumn{3}{c}{\textit{it-sal}} & \multicolumn{3}{c}{\textit{it-sal-RG}} \\ \cmidrule(lr){2-4} \cmidrule(lr){5-7} \cmidrule(lr){8-10} \\
  $\epsilon$ & Stk-A & Stk-FP & Vid-A & Stk-A & Stk-FP & Vid-A & Stk-A & Stk-FP & Vid-A \\ \midrule
  0 		& 74.01	& -  & 80.96 & 74.01 & - 	& 80.96 & 74.01	& -		& 80.96 \\
  0.005 	& 25.74	& 11 & 29.44 & 25.25 & 2.44	& 28.88 & 3.80	& 2.94	& 3.91 \\
  0.01 		& 24.79	& 11 & 28.51 & 23.66 & 2.19	& 26.90 & 1.45	& 2.54	& 1.02 \\
  0.015 	& 23.63	& 11 & 28.20 & 22.08 & 2.23	& 25.53 & 0.78	& 2.37	& 0.57 \\
  0.02 		& 19.97	& 11 & 23.68 & 17.94 & 2.46	& 20.69 & 0.27	& 2.21	& 0.31 \\
  0.025 	& 17.83	& 11 & 21.27 & 15.79 & 2.55	& 18.79 & 0.21	& 2.15	& 0.21 \\
  0.03 		& 15.36	& 11 & 18.55 & 13.31 & 2.66	& 16.15 & 0.10	& 2.11	& 0.10 \\
  0.035 	& 12.90	& 11 & 15.48 & 11.08 & 2.75	& 13.21 & 0.08	& 2.07	& 0.03 \\ \midrule
  \end{tabular}
\end{table}

\subsubsection{Sparsity Results}

The next major result is the sparsity of the attacks. Here, sparsity refers to the number of frames perturbed versus the number of frames in the stack. For an adversarial example to be considered sparsely perturbed, the number of perturbed frames in the stack has to be strictly less than the stack length. Otherwise, the example would be considered densely perturbed. The first result comes from Table \ref{table:acc_results_table}, where the "Stk-FP" columns under each attack variant show the average number of frames perturbed for successful adversarial examples. As expected, the \textit{one-shot} attack yields densely perturbed examples, and both iterative attacks yield sparsely perturbed examples on average. Interestingly, the perturbations are quite sparse, as both iterative variants only require between 2 and 3 frames to be perturbed on average for a successful adversarial example. There also appears to be a relationship between epsilon and average number of frames perturbed. In the non-RG variant the average number of frames perturbed mostly increases with $\epsilon$, while in the RG variant the number of frames perturbed strictly decreases as $\epsilon$ increases. We postulate that this is due to the differing success rate deceleration between the two variants. Since the success rate of the RG variant attack does not change drastically as $\epsilon$ increases, the attack tends to improve the perturbation sparsity of previously successful examples. On the other hand, not only does the non-RG variant improve sparsity on previously successful attacks, but it also achieves many more new successes as $\epsilon$ increases, driving the average frames perturbed on success upwards.


\begin{figure}[h]
  \includegraphics[width=.6\columnwidth]{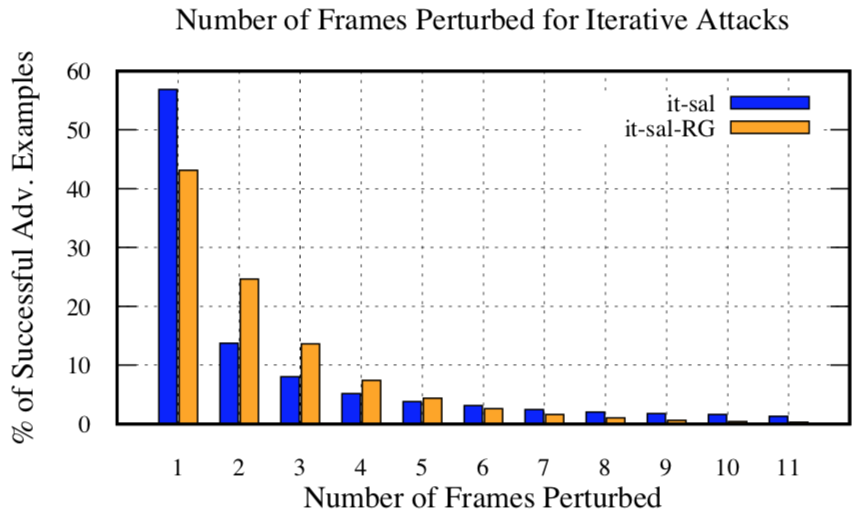}
  \centering
  \caption{Percent of successful adversarial examples versus number of frames perturbed for iterative attack variants. Each bar represents an average of results from all tested epsilons from 0.005 to 0.035.}
  \label{fig:frames_pert}
\end{figure}

Another way to view sparsity results is through histograms of the number of frames perturbed for successful attacks. Fig. \ref{fig:frames_pert} shows the distributions of the number of frames perturbed across all successful adversarial examples for both iterative attacks. Each bar represents the average across all epsilons of attack from 0.005 to 0.035.

The most striking result from this plot is that most successful attacks only perturb a single frame. For the non-RG attack, nearly 60\% of successful attacks require only a single frame perturbation. In the RG attack, over 40\% of successful attacks require a single frame perturbation. However, keep in mind that the RG attack has many more successful examples, so this data does not reflect that the non-RG variant is more effective at perturbing a single frame (in fact, both attacks perturb the first frame exactly the same way). Also, this result shows that the RG attack is more likely to be successful on the subsequent perturbations if the 1st perturbation fails, and is less likely to require more than 6 perturbed frames.

\subsection{Video Level Results}

All results thus far have been at the stack level. However, action recognition and video classifiers ultimately work to classify whole videos, which are potentially comprised of many stacks. As described in \cite{2stream}, an effective way of aggregating stack level predictions into a single video level classification is to average the individual stack predictions. Using this idea, we create a video classifier based only on the temporal stream. Given a whole video, we sample all possible non-overlapping frame stacks. We then classify each stack independently and average the predictions to calculate a single video level prediction. Here, we are able to achieve 80.96\% top-1 test accuracy on split-01 of UCF-101, which is close to the reported 83.7\% temporal-stream-only classifier from the original two-stream paper.

The attacks are straightforward to apply to the video classifier. We attack each stack independently, while maintaining temporal ordering. For the \textit{one-shot} attack, all frames of all stacks are perturbed, always. For both iterative methods, we start by attacking the first stack, then continue to the following stacks as needed. After each stack is perturbed, the video level prediction is measured. If at any point the video is adversarial, the attack stops. This means that not all stacks are attacked as a rule, creating even greater sparsity for the iterative attacks. 




Fig. \ref{fig:video_results_plot} and "Vid-A" columns of Table \ref{table:acc_results_table} show the results of the video attacks. From Fig. \ref{fig:video_results_plot} the video level results are very similar to the stack level results. This indicates that the stack level results are a good indication of an attack's capabilities at the video level. The \textit{one-shot} and \textit{it-saliency} attacks perform similarly to each-other once again, dropping accuracy sharply at $\epsilon=0.005$, then leveling out and never achieving random accuracy. Meanwhile, the \textit{it-saliency-RG} is strictly better than the other two attacks, maintaining a wide margin of performance benefits and achieving random accuracy at $\epsilon=0.015$. Overall, the video level tests show that while these attacks are designed to operate on individual stacks, they can successfully be extended to the video level.

\subsection{Black-box Transferability Results}

To this point, all of the attacks have been under white-box assumptions. We use FlowNet2 as the optical flow algorithm so we can compute gradients through the classifier \textit{and} the optical flow step, back to the video frames themselves. However, it may not be safe to assume the action recognition classifier is using FlowNet2. Rather, the system may be using another algorithm such as TV-L1 or Farneback, where the gradients cannot be computed through the optical flow algorithm. In this setting we test the transferability of adversarial examples created with our white-box model, to action recognition systems using TV-L1 and Farneback algorithms. Here, the MUA is a black-box model that takes a stack of frames and outputs a single prediction.

To test the transferability of adversarial examples created with our white-box method to black-box models, we consider both the \textit{one-shot} and \textit{it-saliency-RG} attacks. For the \textit{one-shot} attack, a densely perturbed stack of frames is generated for the white-box model, then input to the black-box model to test if it is adversarial. For the \textit{it-saliency-RG} attack, the gradients are calculated with the white-box model, but the attack success condition is checked against the black-box model. In other words, the attack iterates until the black-box model misclassifies the stack, even though the perturbations are made with respect to the white-box model. 

\begin{figure}[h]
  \includegraphics[width=.65\textwidth]{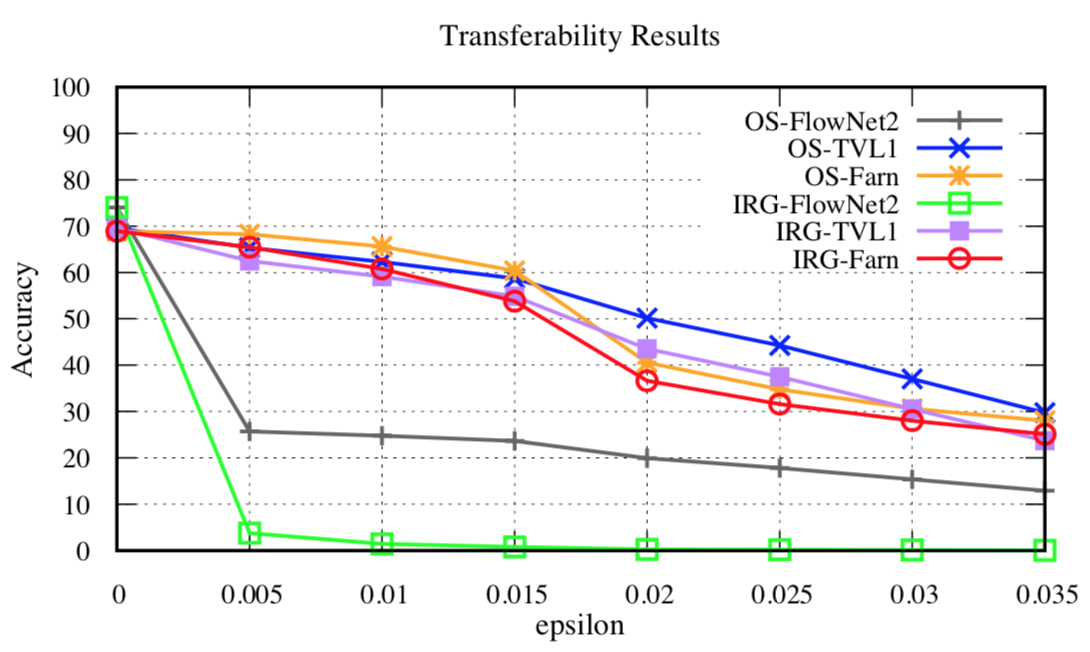}
  \centering
  \caption{This plot shows stack level transferability of attacks by attacking black-box models with examples generated in the white-box setting. The prefix of the label represents the attack variant, OS=one-shot and IRG=it-saliency-RG. The suffix is the system being attacked. The FlowNet2 system is the baseline as it is the white-box system. TVL1 and Farn represent black-box systems with TV-L1 and Farneback optical flow algorithms and CNN trained models, respectively.}
  \label{fig:transferability_results_plot}
\end{figure}

Fig. \ref{fig:transferability_results_plot} and Table \ref{table:transferability_results_table} in the Appendix show the transferability results for the attacks. We immediately see that transferred adversarial examples are not as effective on black-box models (TVL1, Farn) and do not surpass the baseline white-box results of \textit{OS-FlowNet2} or \textit{IRG-FlowNet2}. This is expected because the optical flow algorithms have different properties and the attacks are highly optimized for the white-box models. Also, we see the transferred examples are not very effective at low epsilons but do still significantly decrease accuracy at high epsilons. One interesting result comes when we inspect the difference between TV-L1 and Farneback systems. At $\epsilon < 0.015$ the attacks transfer better to the TV-L1 systems, but at the higher epsilons the attacks transfer better to the Farneback systems. It is unclear why this trend exists and provides an interesting future work. Also, from the previous results we may expect \textit{it-saliency-RG} (IRG) to significantly outperform the \textit{one-shot} (OS) attack in this black-box setting. However, this is not the case in Fig. \ref{fig:transferability_results_plot}. At $\epsilon=0.035$ on the baseline FlowNet2 system IRG outperforms OS by nearly 13\%, on TV-L1 systems IRG only outperforms OS by about 6\%, and on Farneback systems IRG only outperforms OS by about 3\%. This shows that the IRG attack is the most highly optimized for the white-box setting but does not produce more generalized adversarial examples that transfer to a black-box system.  

\section{Conclusion}

Inspired by the recent success of action recognition systems, and the explosion of research on adversarial attack methods, our goal is to develop an attack for action recognition and video classification systems. In this work, we develop an effective attack technique for the widely used optical flow-based classification models in white-box and black-box settings. The attack combines the gradients of a differentiable optical flow calculation algorithm and a convolutional neural network to ultimately perturb the video frames themselves. We show three variants of attack, all of which are capable of significantly degrading classifier accuracy. We also show that we can create sparsely perturbed examples that often only require a single frame perturbation. We also describe a black-box attack that leverages the transferability property of the white-box model to significantly impact a black-box classifier's performance. 

The most pertinent area of future work is to further investigate the transferability of examples. We will work to create examples that transfer better and more reliably, create other variants of black-box attacks, and look to other deep learning optical flow methods that allow for gradient calculations. We will also work to defend models from the attacks described here.


\bibliography{iclr2019_conference}
\bibliographystyle{iclr2019_conference}

\section{Appendix}


\begin{figure}[h]
  \includegraphics[width=\columnwidth]{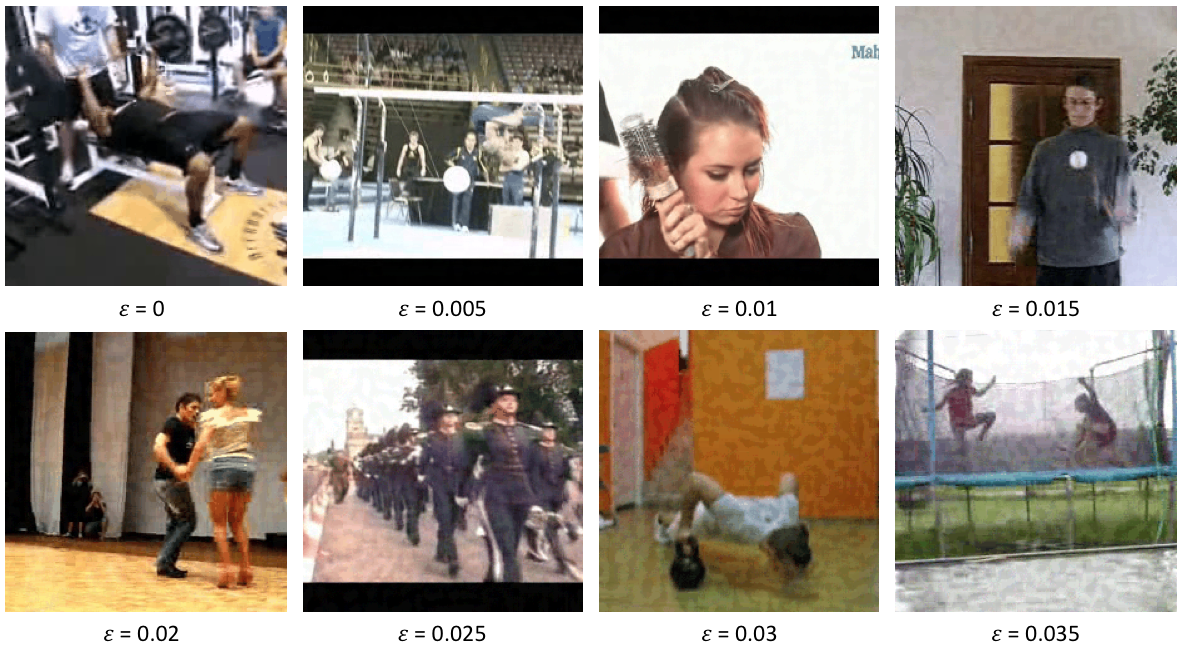}
  \centering
  \caption{Examples of perturbations at each tested epsilon.}
  \label{fig:perturbation_examples}
\end{figure}

Fig. \ref{fig:perturbation_examples} shows examples of individual frames perturbed at each epsilon tested. As expected, as epsilon increases the perturbations becoming more obvious. However, from these still images it is evident that scene complexity plays a role in perceptibility. Although the band marching frame in $\epsilon=0.025$ is perturbed by a stronger adversary, the noise is arguably less perceptible than the dancing frame in the $\epsilon=0.2$ frame.


\begin{table}[h]
  \centering
  \caption{Stack level transferability table.}
  \label{table:transferability_results_table}
  \begin{tabular}{@{}ccccccc@{}}
  & \multicolumn{3}{c}{\textit{one-shot}} & \multicolumn{3}{c}{\textit{it-sal-RG}} \\
  $\epsilon$ 	& FNet2 	& TV-L1 	& Farn 	& FNet2 	& TV-L1 	& Farn 	\\ \midrule
  0 			& 74.01	& 69.61	& 68.94	& 74.01	& 69.61	& 68.94	\\
  0.005 		& 25.74	& 65.34	& 68.26	& 0.03	& 62.49	& 65.48	\\
  0.01  		& 24.79	& 62.32	& 65.62	& 0.01	& 59.08	& 60.77	\\
  0.015  		& 23.63	& 58.66	& 60.34	& $<$0.01	& 54.85	& 53.84	\\
  0.02  		& 19.97	& 50.19	& 40.56	& $<$0.01	& 43.53	& 36.62	\\
  0.025  		& 17.83	& 44.27	& 34.88	& $<$0.01	& 37.53	& 31.62	\\
  0.03  		& 15.36	& 37.04	& 30.64	& $<$0.01	& 30.56	& 28.05	\\
  0.035  		& 12.90	& 29.76	& 28.01	& $<$0.01	& 23.75	& 25.10	\\ \midrule
  \end{tabular}
\end{table}

Table \ref{table:transferability_results_table} is shows the transferability results for the black-box attack and is supplemental to Fig. \ref{fig:transferability_results_plot}. We tested the one-shot and iterative-saliency-RG attacks in this setting and for each attack recorded the accuracy of the Flownet2 MUA (white-box) and the TV-L1 and Farneback MUA's (black-box). Clearly, the attacks are not as powerful in the black-box setting however they do still significantly impact model performance, especially at higher epsilon values.

\end{document}